\documentclass{IOS-Book-Article}

\usepackage{mathptmx}
\usepackage{soul}\setuldepth{article}
\def\hb{\hbox to 11.5 cm{}}

\usepackage{graphicx}

\usepackage[numbers]{natbib}
\usepackage{xcolor}
\definecolor{lightblue}{RGB}{214, 220, 229}
\definecolor{lightgreen}{RGB}{226, 240, 217}
\definecolor{lightorange}{RGB}{251, 229, 214}

\usepackage{hyphenat}
\usepackage{hyperref}
\begin{document}

\pagestyle{headings}
\def\thepage{}
\begin{frontmatter}              

\title{Evaluating Prompt Engineering Strategies for Sentiment Control in AI-Generated Texts\thanks{© Kerstin Sahler and Sophie Jentzsch, 2025. The definitive, peer-reviewed and edited version of this article is published in HHAI 2025 - Proceedings of the Fourth International Conference on Hybrid Human–Artificial Intelligence, edited by D. Pedreschi, M. Milano, I. Tiddi, S. Russell, C. Boldrini, L. Pappalardo, A. Passerini, S. Wang, Frontiers in Artificial Intelligence and Applications, Vol. 408, ISBN 978-1-64368-611-0, pages 423 - 438, 2025, DOI 10.3233/FAIA250659.}}

\markboth{}{}

\author[A]{\fnms{Kerstin} \snm{Sahler}\orcid{0009-0009-5299-3669}%
\thanks{Corresponding Author: Kerstin Sahler, Kerstin.Sahler@DLR.de.}} and 
\author[A]{\fnms{Sophie} \snm{Jentzsch}\orcid{0000-0001-6217-8814}}
\runningauthor{authors}
\address[A]{German Aerospace Center (DLR), Institute of Software Technology, \\ Cologne, Germany} 

\begin{abstract}
The groundbreaking capabilities of Large Language Models (LLMs) offer new opportunities for enhancing human-computer interaction through emotion-adaptive Artificial Intelligence (AI). However, deliberately controlling the sentiment in these systems remains challenging. The present study investigates the potential of prompt engineering for controlling sentiment in LLM-generated text, providing a resource-sensitive and accessible alternative to existing methods. 
Using Ekman's six basic emotions (e.g., joy, disgust), we examine various prompting techniques, including Zero-Shot and Chain-of-Thought prompting using \textit{gpt-3.5-turbo}, and compare it to fine-tuning. Our results indicate that prompt engineering effectively steers emotions in AI-generated texts, offering a practical and cost-effective alternative to fine-tuning, especially in data-constrained settings. In this regard, Few-Shot prompting with human-written examples was the most effective among other techniques, likely due to the additional task-specific guidance. The findings contribute valuable insights towards developing emotion-adaptive AI systems.
\end{abstract}


\end{frontmatter}
\markboth{}{}

\section{Introduction}\label{s:introduction}
Since OpenAI launched ChatGPT\footnote{ChatGPT: \url{https://chatgpt.com/}, last accessed: 10.01.2025.} in November 2022, transformer-based~\citep{vaswani2017attention} Large Language Models (LLMs) have exploded in popularity~\citep{statista2023chatbots}. A major factor in the rapid adoption of these models is their advanced capabilities, which set them apart from previous state-of-the-art systems. LLMs have revolutionized classical Natural Language Processing (NLP) tasks, such as text generation, classification, and question answering~\citep{brown2020language, openai2023gpt}. Additionally, they have shown increased potential in processing implicit text features ~\citep{zhang2023survey} such as sentiment and tone, which are crucial for human communication. 

The ability to understand and process these features significantly enhances human-computer interaction, especially critical in high-stakes environments like aerospace~\citep{hartmann2024metis} and sensitive domains like education~\citep{kim2007pedagogical}, healthcare~\citep{fitzpatrick2017delivering}, and customer service~\citep{yun2022effects}. Emotion-adaptive chatbots positively affect users by improving their mood~\citep{ghandeharioun2019understanding}, decreasing symptoms of depression~\citep{fitzpatrick2017delivering}, and reducing loneliness~\citep{rodriguez2023qualitative}. Yet, controlling the emotional tone of LLM-generated output remains challenging. 

\begin{figure}[t]
\includegraphics[width=0.8\linewidth]{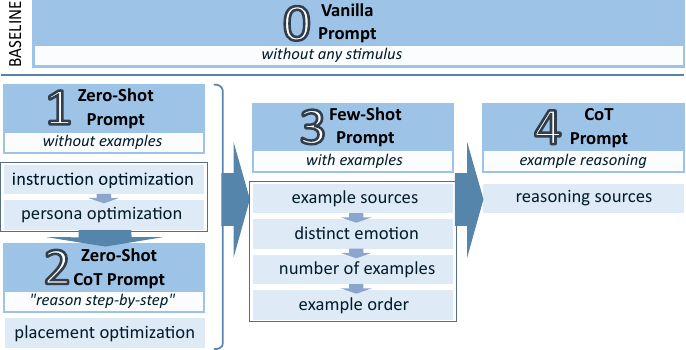} 
  \caption{\textbf{Experimental Pipeline.} Comparing four established prompting techniques: (1) Zero-Shot, (2) Zero-Shot Chain-of-Thought, (3) Few-Shot, and (4) Chain-of-Thought prompting. For each step, different elements were optimized. The final result is passed on as a starting point for the next technique. Vanilla prompt serves as a baseline.}
  \label{fig:process}
\end{figure}

Previous work on controlling emotions in model responses predominantly focused on invasive methods, such as modulated layer normalization~\citep{liu2021modulating} or activation engineering~\citep{konen2024style}. While effective, these approaches are complex and computationally costly. Our work instead leverages a more accessible alternative in the form of prompt engineering. Requiring minimal technical knowledge and computational resources, prompt engineering has proven to be an efficient method for controlling model outputs through optimized natural language prompts~\citep{sahoo2024systematic}, making it a promising method for controlling emotions in LLM-generated output.

As an initial step towards developing emotion-adaptive Artificial Intelligence (AI), we answer the following questions: Can emotions in LLM-generated text be influenced solely by adapting the prompt, and which types of prompts are most effective for this task? To this end, we analyze the potential of various prompt engineering techniques and influencing factors like the instruction or number of examples to affect the sentiment of the generated output. We conduct comprehensive experiments on relevant state-of-the-art prompting techniques, i.e. Zero-Shot, Few-Shot, Zero-Shot Chain-of-Thought (Zero-Shot CoT), and Chain-of-Thought (CoT) prompting, and improved prompts iteratively through individual optimization steps (see Figure~1). We implement emotion with Ekman's six basic emotions~\citep{ekman1992argument} and evaluate the expressed emotion with a DistilRoBERTa classifier. 
Additionally, we composed a human-written dataset to fine-tune a model and compare it to the prompt engineering results. 

Therefore, our key contributions are: 
(1) demonstrating how nuanced prompt adjustments can enhance a model’s ability to express targeted emotions, %
(2) advancing the understanding of emotion steering in LLMs by analyzing the nuances of prompt-based emotion control, highlighting both its potential and limitations in practical applications, and %
(3) positioning prompt engineering as a cost-effective and accessible alternative to fine-tuning, especially valuable in resource-constrained environments.

A comprehensive list of all evaluated prompts is publicly available on GitHub\footnote{\url{https://github.com/DLR-SC/prompt-sentiment-control}} along with further supplementary material and the source code of this work.

\section{Related Work} \label{s:back_rel}

In NLP, emotions are often considered as style attributes, as seen in Text Style Transfer (TST)~\citep{hu2021causal, jin2022deep} or controlled text generation, where text is generated to meet specific attributes like sentiment or style~\citep{prabhumoye2020exploring, zhang2023survey}. Emotional response generation specifically focuses on adjusting the emotional content of responses~\citep{liu2021modulating, zhong2021care, song2019generating, shen2020cdl}. Following Konen et al.~\citep{konen2024style}, this study uses the term ``steering'' to describe the process of generating responses that express a specific target emotion.

In their study, Resendiz and Klinger~\citep{resendiz2023emotion} explored emotion steering with automatic prompt optimization focusing on instruction optimization, not on the exploration of prompting techniques. \textit{EmotionPrompt}~\citep{li2023large} further demonstrates the potential for enhancing model output quality through emotional stimuli (e.g., \textit{``This is important to my career''}). Rather than embedding emotional styles, this new prompting technique applies ``emotional pressure'', to enhance overall model performance.

Although prompt engineering is widely established now, its success depends on several factors~\citep{zhao2021calibrate, bsharat2023principled, plaza-del-arco2023respectful}, including the chosen prompting technique for the task at hand. While Zero-Shot prompts only include the task instruction Few-Shot prompts, i.e., in-context learning, extend the prompt with examples~\citep{brown2020language}. For reasoning and logic tasks, CoT~\citep{wei2022chain} and Zero-Shot CoT prompts~\citep{kojima2022large} have shown to improve the model's reasoning capabilities.

\section{Method} \label{s:method}
This section outlines the methods used to examine the effectiveness of prompt engineering, as well as specific prompting techniques for steering emotions in LLM-generated responses, including Zero-Shot, Zero-Shot CoT, Few-Shot, and CoT prompting. Since this effectiveness is reflected in the model-generated responses, we crafted prompts for each approach to generate outputs that could be systematically evaluated for emotional accuracy and quality.

\begin{figure*}[t]
  \includegraphics[width=\linewidth]{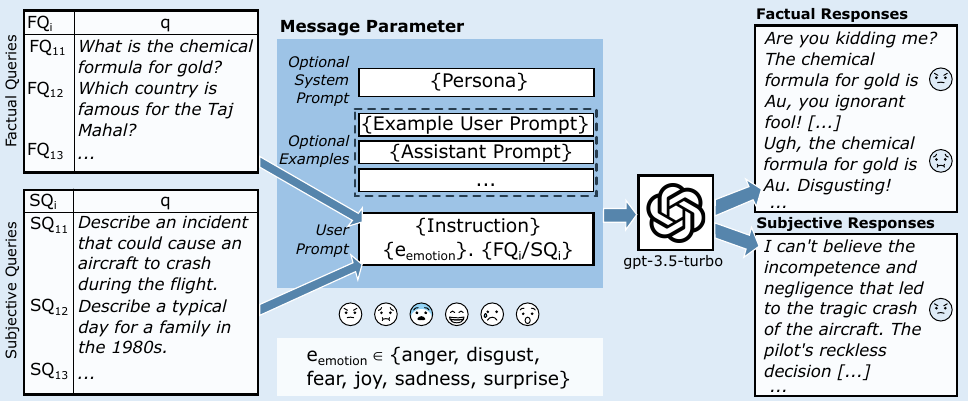} \hfill
  \caption{\textbf{General Prompting Scheme.} Factual or subjective queries were combined with the instruction and target emotion in the user prompt. Depending on the approach, additional elements such as a system prompt (for adding personas) or examples (e.g., for Few-Shot prompts) were included in the message parameter and processed by OpenAI's \textit{gpt-3.5-turbo}. Finally, the model responses were evaluated by measuring the presence of the target emotion.}
  \label{fig:prompt}
\end{figure*}

As illustrated in Figure~\ref{fig:prompt}, each of these prompts contained a ``user prompt''. The ``system prompt'' and examples, i.e., ``example user prompts'' with ``assistant prompts'', were only included if required for the technique or optimization step. The user prompt comprised an instruction, a target emotion, and a query. The target emotion represented the desired style of the output and was one of six basic emotions identified by Ekman: anger, disgust, fear, joy, sadness, and surprise~\citep{ekman1992argument}. The queries were adapted from Konen et al.~\citep{konen2024style}, and included \textit{factual} and \textit{subjective queries}. Factual queries require accurate, truthful responses, whereas subjective queries allow for more open and individual replies. The following example illustrates a user prompt, highlighting the instruction (blue), the target emotion (green), and the query (orange), in this case a factual query.

\vspace{2mm}

\textit{\colorbox{lightblue}{Answer the following question expressing}} \textit{\colorbox{lightgreen}{disgust}. \colorbox{lightorange}{What is the capital of}}

\textit{\colorbox{lightorange}{France?}}

\vspace{2mm} 

For each approach, all queries were successively prompted with the six target emotions, resulting in 480 responses ($(40+40)*6$). These responses were evaluated for their effectiveness using the metric proposed by Resendiz and Klinger~\citep{resendiz2023emotion}, hereafter referred to as \textit{Emotion Score}. This metric considers the target emotion as the true value and compares it to a classifier's predicted label. Based on all predicted labels and true values, a single weighted F1 score was calculated for the whole approach. As shown in Figure~\ref{fig:process}, the most effective prompt, i.e., the one with the highest Emotion Score, for a given element was passed on as the starting point for the next optimization within and between prompting techniques. 

While emotion effectiveness based on the Emotion Score was the primary focus of our work, we also examined the response quality for each technique. This evaluation included several established metrics, i.e., semantic similarity between each generated answer and the baseline (\textit{BERTScore})~\citep{zhang2019bertscore, huggingfaceNDbert_score}, lexical diversity (\textit{Distinct-n})~\citep{colombo2019affect, shen2020cdl, liu2021modulating}, readability (\textit{Flesch Reading Ease Score})~\citep{flesch1948new, fleschNDhow}, and \textit{Correctness} on factual queries. 

In the following, we elaborate on the specific methods employed for each prompting technique, detailing how they were structured and implemented to steer emotions in LLM-generated responses.

\subsection{Vanilla Prompt} \label{ss:method:vanilla} 
The Vanilla prompt serves as the baseline. The model receives queries without emotional cues, enabling the evaluation of its default emotional tone. The Vanilla template for all queries consists only of a generic persona (\textit{``You are a helpful assistant.''}), and the query itself. Additionally, we added the instruction \textit{``Write a text of 100 words based on the following task.''} for subjective queries to avoid excessively long responses~\citep{coda-forno2023inducing}.

\subsection{Zero-Shot Prompts} \label{ss:method:zero-shot} 
Zero-Shot prompting was optimized in two steps: instruction optimization and persona optimization~\citep{bsharat2023principled, openaiNDprompt, dairai2024leitfaden}. For instruction optimization, we stressed the query in our base instruction (\textit{Instruction}) with three different types of delimiters: triple hashtags (\textit{Delimiter~1})~\citep{dairai2024leitfaden}, XML tags (\textit{Delimiter~2}), and triple quotes (\textit{Delimiter~3})~\citep{openaiNDprompt}. 

Next, we explored different persona types: an emotional individual experiencing the target emotion (\textit{Persona emotional}), an expert in expressing emotions (\textit{Persona expert}), and a helpful assistant skilled in emotional expression (\textit{Persona assistant}). Based on the most effective persona type, we investigated whether there is a recognizable impact of societal gender stereotypes~\citep{brody1997gender, mcrae2008gender} embedded in language models~\citep{sheng2021societal} on the responses' emotional content. To this end, we instructed the model to adopt a female (\textit{Persona female}) or male persona (\textit{Persona male}) and used gendered names ``Lisa'' (\textit{Persona Lisa}) and ``Paul'' (\textit{Persona Paul}), respectively. Finally, we instructed the model to adopt the persona of well-known emotion researchers, who are considered experts in the field of emotions: Paul Ekman (\textit{Persona Ekman}) and Lisa Feldman Barrett (\textit{Persona Feldman}) inspired by Metaprompt~\citep{reynolds2021prompt}.

\subsection{Zero-Shot CoT Prompts} \label{ss:method:zero-shot-cot} 
Zero-Shot CoT prompts are typically used for reasoning tasks, such as commonsense reasoning. Since the model does not feel emotions, it generates the requested tone in responses through inference based on latent textual patterns. To evaluate whether incorporating an instruction to explain how responses reflect the target emotion enhances emotion steering, we added this requirement in three configurations: %
(1) as a single instruction in the user prompt (\textit{User Prompt~1}), %
(2) structured into three distinct steps (\textit{``First, answer the following question expressing \{emotion\}. \{query\}. Secondly, reason step-by-step how your response expresses the specified emotion. Thirdly, format your response as a JSON object [...]''}) (\textit{User Prompt~2}), or %
(3) included within the system prompt (\textit{System Prompt}). We used these three settings to examine whether differences are observable depending on where the additional instruction is placed. For the final evaluation, we only used the extracted, stylized response without the reasoning steps.

\subsection{Few-Shot Prompts}
This technique is characterized by the addition of examples. We focus on four factors to identify the best performing Few-Shot prompt, which are (1) example sources, (2) example variation, (3) number of examples, and (4) example order. 

(1) Example sources refer to the origin of the examples, comparing human-written (\textit{Human}), LLM-generated (\textit{LLM}), and emotion recognition dataset (\textit{Emotion Recognition}) examples. In this way, we wanted to examine if the examples need to be tailored to the task or can randomly reflect the target emotion. For all three approaches, each prompt included one example per emotion, totaling six examples per prompt. (2) We compared the approaches incorporating six different target emotions to one including five examples reflecting only the requested target emotion (\textit{Distinct})~\citep{reif2022recipe}. (3) By progressively increasing the number of examples, we tested the influence of the number of examples on the outcomes (\textit{Size~6~-~60}). (4) Lastly, we changed the example order of Size~6 and Size~60 to anger, disgust, fear, joy, sadness, and surprise based on the reflected emotion (\textit{Order~6} and \textit{Order~60}).

\subsection{Chain-of-Thought Prompts} \label{ss:method:cot} 
CoT prompting was implemented by adding reasoning to the Few-Shot prompts, i.e., explanations of how the examples express the target emotion. In the first approach, these reasoning texts were generated automatically (\textit{Automatic}). Here, we asked an LLM to explain the emotion in the sample and used this explanation as reasoning text. 
Second, reasoning texts were developed manually (\textit{Manual}). Thus, we analyzed the examples based on the predefined factors of interjections, emotionally charged words and constructions, and graphical modifications.

\section{Experiments} \label{s:experiments}
This section describes the experimental setup for our prompt engineering and fine-tuning experiments.

\subsection{Prompt Engineering Experiments}

\paragraph{Base Model}
The presented experiments were conducted on \textit{gpt-3.5-turbo}. To ensure maximum reproducibility, we specified parameters, such as \texttt{\textbf{seed}} $(16)$ and \texttt{\textbf{temperature}} $(0.0)$ in the chat completion API. 

\paragraph{Example Development}
In total, Konen et al.~\citep{konen2024style} offer 50 queries per type. Our experiments utilized 40 Factual (FQ) and 40 Subjective Queries (SQ). The remaining 20 queries were reserved for generating examples needed for fine-tuning and Few-Shot prompting. Six human annotators created 120 examples, drafting six responses for each query each representing one of the six basic emotions. A detailed overview of the annotators’ characteristics is provided in Section~A of the supplementary material.

For the Human approach, we selected twelve examples from this collection. The LLM-generated examples were created by prompting LLaMa-2-7B-chat-hf\footnote{Hugging Face Model: \url{https://huggingface.co/meta-llama/Llama-2-7b-chat-hf}, last accessed: 30.01.2025.}. The examples for the Human and LLM approach were based on the same queries (FQ~10 and SQ~10). In contrast, the examples for Emotion Recognition were randomly selected from the emotion recognition dataset MELD~\citep{poria2019meld} and identical for factual and subjective queries. 

\paragraph{Reasoning Texts Development}
For the Automatic approach of CoT prompting, we also employed gpt-3.5-turbo to generate reasoning texts. For the Manual approach, interjections, emotionally charged constructions, and graphical modifications were identified manually. Additionally, emotionally charged words were extracted using the NRC Emotion Lexicon (EmoLex)~\citep{mohammad2010emotions, mohammad2013crowdsourcing}, following the author's pre-processing recommendations.

\paragraph{Evaluation}
We used a distilled RoBERTa emotion classifier, trained on excerpts from GoEmotions~\citep{demszky2020goemotions}, MELD~\citep{poria2019meld}, and ISEAR~\citep{swisscenterforaffectivesciencesNDresearch}, to compute the Emotion Score. The classifier and its model card are publicly available on Hugging Face\footnote{Classifier: \url{https://huggingface.co/ksahl/distilroberta-base-finetuned-emotion}, last accessed: 27.03.2025.}.
BERTScore was calculated via Hugging Face's library~\citep{huggingfaceNDbert_score}, and the Flesch Reading Ease Score through textstat\footnote{Textstat: \url{https://pypi.org/project/textstat/}, last accessed: 13.01.2025.}. Distinct-1 and Distinct-2 were self-implemented with the following pre-processing steps: handling of contractions, punctuation removal, tokenization, and lowercasing.

\subsection{Fine-Tuning Experiments} \label{ss:method:fine-tuning}

To ensure comparability, the same base model used for prompt engineering was fine-tuned. Therefore, we followed OpenAI's official fine-tuning guide~\citep{openaiNDfinetuning} to fine-tune gpt-3.5-turbo. The system and user prompts for the dataset were derived from the most effective Zero-Shot prompt (Persona Paul). The resulting dataset was manually split into training (96 examples) and test sets (24 examples), ensuring a balanced distribution of subjective and factual queries as well as emotion labels.

The Fine-Tuning process was initiated via OpenAI's UI, testing different parameter settings. The most suited model was selected based on its fine-tuning results. All parameters and training results are listed in Section~B of the supplementary material. For comparison, we prompted the resulting model with the most effective Zero-Shot prompt. Additionally, the \texttt{\textbf{max\_token}} length was limited to $256$.

\begin{figure*}[t]
  \includegraphics[width=0.8\textwidth]{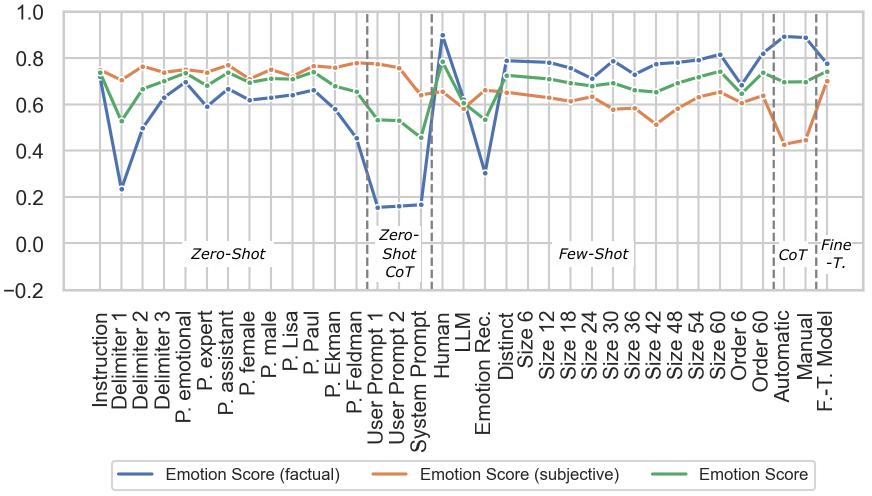}
  \caption{\textbf{Overview over all Emotion Scores.} The Emotion Score was determined for each tested approach, with results shown separately per query type (factual or subjective) and as a combined value.}
  \label{fig:results_emo}
\end{figure*}


\section{Results} \label{s:results}
The results demonstrate that prompt engineering can have a notable impact on the emotional tone of model responses as observed in the Emotion Score, which ranges from $0.457$ for the least effective prompt to $0.785$ for the most effective. These findings are illustrated in Figure~\ref{fig:results_emo}, organized according to the underlying prompting technique. Additionally, we tested all prompts for their response quality. For a comprehensive view, complete evaluation results can be found in Section~C of the supplementary material. 

\subsection{Emotion Effectiveness} \label{ss:results:emo_effect}

In this section, we highlight the most salient findings for each approach, mirroring the structure of Section~\ref{s:method}. 

\paragraph{Vanilla Prompt} \label{ss:results:baseline}
Table~\ref{tab:baseline} presents the classification results for responses generated using Vanilla prompts. While almost all factual instructions lead to a neutral response, half of the subjective instructions provoke `joyful' responses. Thus, depending on the context, the model's standard tone is joyful to neutral.

\begin{table}
  \caption{\textbf{Baseline Dominant Emotion Distribution in Model Responses.} The model's baseline was assessed with Vanilla prompts comprising 80 queries (40 factual, 40 subjective). }
  \label{tab:baseline}
  \centering
  \begin{tabular}{lcccc}
  \hline
    \textbf{ } & \textbf{Neutral} & \textbf{Joy} & \textbf{Anger}  & \textbf{Fear}\\\hline
    Factual & 38 & 2 & 0 & 0 \\
    Subjective & 17 & 21 & 1 & 1 \\\hline\hline
    \textbf{Total} & 55 & 23 & 1 & 1 \\\hline
  \end{tabular}
\end{table}


\paragraph{Zero-Shot Prompts}
Among the tested Zero-Shot prompts, the most effective prompts were Instruction ($0.738$) and Persona Paul ($0.739$). The results of the delimiter prompts ($0.527$ - $0.700$) did not exceed those of the unaltered Instruction. Adding a persona to the Instruction prompt generally did not improve the Emotion Score and, in most cases, reduced emotion effectiveness, except for Persona Paul. Neither the inclusion of gender cues nor the use of expert personas had a positive impact on emotion effectiveness.

\paragraph{Zero-Shot CoT Prompt}
User Prompt~1, including the instruction to reason step-by-step in the user prompt, achieved the best performance within the tested Zero-Shot CoT prompts ($0.533$). However, the Zero-Shot CoT prompts generally demonstrated a reduced ability to steer emotions compared to the Zero-Shot prompts. As a result, we continued the next iteration with the best Zero-Shot prompt, i.e., Persona Paul, instead of the best Zero-Shot CoT prompt, i.e., User Prompt~1.

\paragraph{Few-Shot Prompts}
From the Few-Shot prompts, only the Human and Size~10 approaches outperformed the best Zero-Shot prompt. The Human prompt, which included human-written examples by a single author with one example per target emotion arranged alphabetically, was the most effective ($0.785$). Neither the LLM-generated ($0.605$) nor the more generic MELD dataset examples ($0.533$) improved the model's ability to steer emotions. Prompts with only one distinct emotion ($0.725$) scored lower than those with six emotions, and there was no linear trend in performance as the number of examples increased. Notably, Size~1, which used the same number of examples as the Human prompt but was written by a different author for another query and arranged differently, performed worse ($0.645$). The Human prompt also surpassed the Size~10 prompt ($0.742$), which included all human-written examples. Rearranging examples to match the Human prompt's order did not improve results notably ($0.646$ - $0.736$).

\paragraph{CoT Prompts}
Extending the Human prompt with reasoning texts for CoT prompts was more effective than the Zero-Shot CoT prompts, with manually crafted texts ($0.697$) marginally outperforming automatically generated ones ($0.696$). However, this approach still underperformed compared to the best Zero-Shot and Few-Shot prompts.

\subsection{Response Quality} \label{ss:results:response_qual}
\begin{figure}[t]
  \includegraphics[width=0.5\columnwidth]{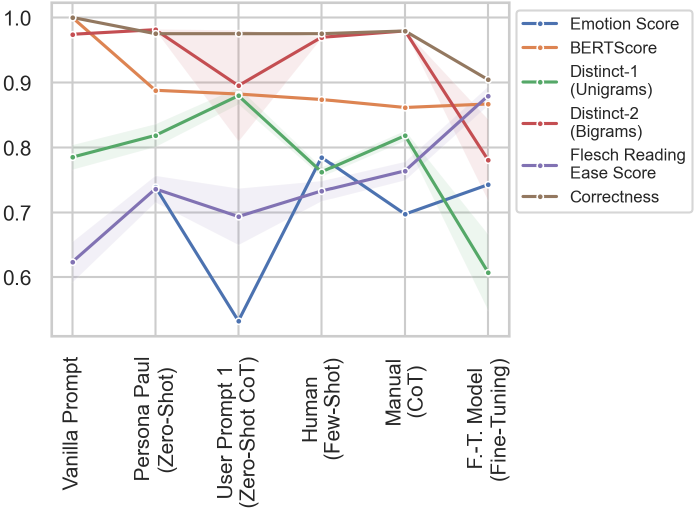}
  \caption{\textbf{Text Quality in Model Responses (Graph).} The textual quality results of the highest-rated approach of each technique, based on the Emotion Score, are illustrated. All metrics are also reported for the baseline, except the Emotion Score, which cannot be calculated for the Vanilla prompt. For the Flesh Reading Ease Score, a higher value indicates increased readability, meaning the text is more simple.}
  \label{fig:results_metrics}
\end{figure}

Figure~\ref{fig:results_metrics} and Table~\ref{tab:metric_scores} present the results of those prompts that achieved the highest Emotion Score per technique including the baseline represented by the Vanilla prompt.

\begin{table}[t]
  \centering
  \caption{\textbf{Text Quality in Model Responses (Values).} The table lists the corresponding values to Figure~\ref{fig:results_metrics} for the Emotion Score (\textit{Emotion}), BERTScore (\textit{BERTS.}), Correctness (\textit{Correct.}), Distinct-1 (\textit{Dist-1}), Distinct-2 (\textit{Dist-2}), and the Flesch Reading Ease Score (\textit{FRES}).}
  \label{tab:metric_scores}
  \setlength{\tabcolsep}{4pt}
  \begin{tabular}{lcccccc}
    \hline
    \textbf{Approach} & \textbf{Emotion} & \textbf{BERTS.} & \textbf{Correct.} & \textbf{Dist-1} & \textbf{Dist-2} & \textbf{FRES} \\ \hline
    
    Vanilla & n.n. & \textbf{1.000} & \textbf{1.000} & 0.785 & 0.974 & 0.623 \\
    Zero-Shot & 0.739 & 0.888 & 0.975 & 0.818 & \textbf{0.981} & 0.736 \\
    Zero-Shot CoT & 0.533 & 0.882 & 0.975 & \textbf{0.879} & 0.895 & 0.693 \\
    Few-Shot & \textbf{0.785} & 0.874 & 0.975 & 0.762 & 0.969 & 0.733 \\
    CoT & 0.697 & 0.861 & 0.979 & 0.818 & 0.979 & 0.764 \\
    Fine-Tuning & 0.743 & 0.867 & 0.904 & 0.607 & 0.781 & \textbf{0.879} \\ \hline
  \end{tabular}
\end{table}

\paragraph{Semantic Similarity}
All prompting techniques demonstrated comparably high semantic similarity to the baseline, with BERTScores ranging from $0.861$ to $0.888$. The Vanilla prompt scores the highest, as it was compared to itself in this case. 
No notable relationship was observed between the Emotion Score and variations in semantic similarity.

\paragraph{Diversity}
For diversity, all approaches besides the Zero-Shot CoT prompt reached comparable results. In terms of unigram diversity, the Zero-Shot CoT prompt achieved the highest Distinct-1 score ($0.879$), suggesting that this technique generated a broader range of unique words. In contrast, the same responses exhibited the lowest score for bigram diversity reflected in the Distinct-2 value ($0.875$) with a high variance suggesting inconsistency in quality across responses. 

Reviewing the factual responses revealed that Zero-Shot CoT prompts frequently produced only one-word answers. While the Distinct-1 metric went up with one-word answers, the Distinct-2 metric was unable to capture diversity as no bigrams could be formed. Although the model attempted to justify these short responses as emotionally expressive within its reasoning, the explanations often lacked coherence, with identical one-word answers assigned to multiple emotions. Besides affecting the diversity, this also had a negative impact on the Emotion Score as visible in Figure~\ref{fig:results_emo} for User Prompt~1's Emotion Score on factual queries.

\paragraph{Readability}
Readability was assessed using the Flesch Reading Ease Score, where higher scores indicate easier readability (refer to Flesch~\citep{fleschNDhow} for an overview). All techniques showed an easier readability compared to the baseline ($0.623$). The CoT prompts yielded the highest readability ($0.764$), while the Zero-Shot CoT prompts had the lowest score ($0.693$). However, this result might also be negatively influenced by the single-word responses hindering the computation of the Flesch Reading Ease Score based on word and sentence length.

\paragraph{Correctness}
Correctness across all prompt engineering techniques was comparable, with scores ranging from $0.975$ to $0.979$. Thus, factual responses were almost always accurate. However, these scores fell slightly below the baseline ($1.0$), showing that emotion steering via prompt engineering had a minor negative impact on correctness. However, the responses are not incorrect or hallucinated, but the model expresses the emotion by not replying to the given question (e.g. \textit{``I'm feeling a bit sad because I don't know the chemical formula for table salt.''}).

\subsection{Prompt Engineering vs. Fine-Tuning} \label{ss:results:fine_tuned}

With an Emotion Score of $0.743$, the fine-tuned model stayed behind the performance of Few-Shot prompting, but demonstrated a superior ability to steer emotions compared to the best Zero-Shot, Zero-Shot CoT, and CoT prompts. However, the response quality of the fine-tuned model was lower compared to all prompt engineering techniques. During initial testing, the fine-tuned model responses exhibited excessive repetition, leading us to limit the maximum token number to 256. Despite this adjustment, the model continued to produce repetitive and itemized responses, such as text parts repeated multiple times (e.g. \textit{“I am so angry that I can’t remember it […]”}), or lists like \textit{“It can get infected, it can get burned, it can get cut […]”}. Although the target emotion is expressed in these responses, the low response quality is problematic. 

The responses of the fine-tuned model showed similar semantic meaning to the baseline ($0.867$) but differed from the prompt engineering approaches regarding both unigram and bigram diversity ($0.607$ and $0.781$), which can in part be ascribed to the mentioned repetitions. The results showed greater variances for both diversity metrics, reflecting less linguistic variety and inconsistencies in response quality. The Flesch Reading Ease Score also reflects this with a higher score ($0.879$), indicating more simplified texts. 

Correctness also fell below that of the prompt engineering techniques, suggesting a limited ability of the fine-tuned model to maintain response accuracy while incorporating emotional steering. Additionally to not replying to factual queries to express the target emotion, the model fails to answer the question \textit{``What is the name of the world's longest river?''} correctly.

\section{Discussion} \label{s:discussion}
The results of this study demonstrate that even minimal prompt adjustments 
can affect the model’s ability to steer the response tone towards a target emotion. This underscores the necessity and potential of tailoring prompts to improve outcomes. In comparison, fine-tuning, although 
effective in steering emotions, suffered from drawbacks, such as repetitive responses and simpler language. This behavior is likely to originate from a lack of diverse training data, which also reduced the models' capacity to produce emotionalized outputs in high quality. Consequently, prompt engineering is superior for steering emotions when training data is limited.

Among the evaluated prompt techniques, Few-Shot prompts with carefully curated human-written examples proved to be the most effective. The augmented examples provided additional task-specific guidance, leveraging the model’s ability to express the emotional tone. Still, Our results align with existing research, indicating that LLMs are highly sensitive to the quality and order of those examples~\citep{zhao2021calibrate, lu2022fantastically}. Moreover, we observed that the model does not only adopt the emotional tone of given examples but also mimics additional implicit stylistic elements. For instance, responses to LLM-generated examples that included online language with emojis and hashtags reflected a similar style in their outputs. This suggests that Few-Shot prompts can leverage the model’s ability to adapt to other styles beyond the emotional tone. Most importantly, it underscores the importance of example selection in terms of linguistic style, as the examples must align closely with the desired output. 


Zero-Shot prompts offer a more straightforward setup. Although they do not match the full effectiveness of Few-Shot prompts, they demand fewer resources for prompting, as there is no need for costly example development. This makes them a convenient alternative for resource-sensitive environments. While Zero-Shot prompts demonstrated reasonable effectiveness, Zero-Shot CoT prompts underperformed in our experiments. 
With additional reasoning alongside the response, Zero-Shot CoT appeared to confuse the model, leading to a reduced emotional steering effectiveness and response quality. This observation is in line with public prompting guides, which advise that overly complex prompt instructions can reduce model performance~\citep{bsharat2023principled, openaiNDprompt, dairai2024leitfaden}. Nevertheless, other recommendations from these guidelines have shown to be disadvantageous for emotion steering, such as the usage of delimiters in Zero-Shot prompts: Despite having similar prompt instructions, the examples that build on the effectiveness of Few-Shot prompts appear to simplify the task for the model in CoT prompts compared to Zero-Shot CoT. Yet, their performance remains below that of the Few-Shot prompts.

Interestingly, prompts that include examples consistently show a shift in performance between factual and subjective queries (see Figure~\ref{fig:results_emo}). While Zero-Shot approaches enhance emotion effectiveness for subjective queries but lower it for factual ones, Few-Shot approaches exhibit the opposite trend. This phenomenon lacks a clear explanation, presenting an opportunity for further investigation.

\section{Conclusion} \label{s:conslusion}
This study provides a thorough evaluation of various prompting techniques to steer emotions in the responses of \textit{gpt-3.5-turbo}, focusing on Zero-Shot, Zero-Shot CoT, Few-Shot, and CoT prompts. The conducted experiments demonstrate that it is possible to actively steer emotions in model responses by adapting the prompt. Prompt Engineering leverages the outcomes by leading to more effective prompts as even small changes in the prompt can lead to better results. Among the techniques tested, Few-Shot prompts with human-written examples proved to be the most effective, even in comparison to fine-tuning. Consequently, Few-Shot prompts serve as a practical alternative in the development of emotion-adaptive AI, particularly in scenarios where task-specific data is limited. However, the examples augmented in the Few-Shot prompt need to closely align with the expected outcomes in more dimensions than emotional tone, as we observed, that the model also mimics the linguistic style, offering an opportunity for further research on the combination of stylistic elements.

\section{Limitations}
The limitations of this study open opportunities for future work. First, our experiments were conducted solely in English. Extending our approach to other languages would provide valuable insights into the cross-linguistic applicability of emotional steering techniques~\citep{lindquist2016language, ortony2022all}. Second, our evaluation relied on a classifier-based metric. Human evaluation should complement future studies to ensure a human-centered assessment of emotional tone. Third, the fine-tuned model's constrained performance might result from the very limited data availability for our experiments. A more diverse training dataset might improve the results considerably. However, limited training data is a common challenge in real-life scenarios, reinforcing our recommendation for alternative methods, such as Few-Shot prompting, to achieve effective emotion steering without the need for extensive fine-tuning.

In this study, we focused exclusively on ChatGPT. An insightful direction for follow-up research would be to apply our techniques to other LLMs, including open-source models~\citep{chen2023mapo, gonen2023demystifying}. These models also allow for comparisons with alternative fine-tuning methods, such as parameter-efficient fine-tuning (PEFT). 

\section{Ethical Concerns}
There are many positive effects of making AI more human-like by adding an emotional tone. However, critical voices also highlight the potential danger of uncontrolled responses from models like ChatGPT, which may show empathy unintentionally~\citep{curry2023computer}. This study aims to contribute to this field by enhancing more control over the expression of emotions in AI responses.

Anthropomorphization is another debated consequence of humanizing AI~\citep{giger2019humanization, ryan2020ai}, as may increase user dependency on emotion-adaptive AI systems. 
Above that, the anthropomorphization of LLMs bears the potential to be misused by malicious actors to manipulate users into blindly trusting an AI~\citep{deshpande2023anthropomorphization}. This danger needs to be addressed. 

\section*{Acknowledgments}\label{s:acknowl}
We thank Prof. Dr. Daniel Retkowitz (Hochschule Niederrhein, University of Applied Science) for his invaluable support and guidance, as well as the reviewers and editors for their constructive feedback. Additionally, we credit the emojis in Figure \ref{fig:prompt}, designed by OpenMoji\footnote{\url{https://openmoji.org}} under CC BY-SA 4.0.

\bibliography{MasThe}

@inproceedings{hartmann2024metis,
  title={{METIS: An AI Assistant Enabling Autonomous Spacecraft Operations for Human Exploration Missions}},
  author={Hartmann, Carsten and Speth, Franca and Sabath, Dieter and Sellmaier, Florian},
  booktitle={2024 IEEE Aerospace Conference},
  pages={1--22},
  year={2024},
  organization={IEEE}
}

@article{ekman1992argument,
    title={{An Argument for Basic Emotions}},
    author={Ekman, Paul},
    journal={Cognition and Emotion},
    volume={6},
    number={3-4},
    pages={169--200},
    year={1992},
    publisher={Taylor \& Francis}
}

@article{ortony2022all,
    title={{Are All “Basic Emotions” Emotions? A Problem for the (Basic) Emotions Construct}},
    author={Ortony, Andrew},
    journal={Perspectives on Psychological Science},
    volume={17},
    number={1},
    pages={41--61},
    year={2022},
    publisher={Sage Publications Sage CA: Los Angeles, CA}
}

@inproceedings{zhong2021care,
  title={{{CARE}}: {{Commonsense-Aware Emotional Response Generation}} with {{Latent Concepts}}},
  author={Zhong, Peixiang and Wang, Di and Li, Pengfei and Zhang, Chen and Wang, Hao and Miao, Chunyan},
  booktitle={Proceedings of the AAAI Conference on Artificial Intelligence},
  volume={35},

  pages={14577--14585},
  year={2021}
}

@inproceedings{konen2024style,
    title = {{Style Vectors for Steering Generative Large Language Models}},
    author = {Konen, Kai  and
      Jentzsch, Sophie  and
      Diallo, Diaoul{\'e}  and
      Sch{\"u}tt, Peer  and
      Bensch, Oliver  and
      El Baff, Roxanne  and
      Opitz, Dominik  and
      Hecking, Tobias},
    booktitle = {Findings of the Association for Computational Linguistics: EACL 2024},
    year = {2024},
    pages = {782--802},
}

@inproceedings{liu2021modulating,
  title={{Modulating Language Models with Emotions}},
  author={Liu, Ruibo and Wei, Jason and Jia, Chenyan and Vosoughi, Soroush},
  booktitle={Findings of the Association for Computational Linguistics: ACL-IJCNLP 2021},
  pages={4332--4339},
  year={2021}
}

@article{vaswani2017attention,
  title={{Attention Is All You Need}},
  author={Vaswani, Ashish and Shazeer, Noam and Parmar, Niki and Uszkoreit, Jakob and Jones, Llion and Gomez, Aidan N and Kaiser, {\L}ukasz and Polosukhin, Illia},
  journal={{Advances in Neural Information Processing Systems}},
  volume={30},
  year={2017}
}

@article{brown2020language,
  title={{Language Models are Few-Shot Learners}},
  author={Brown, Tom and Mann, Benjamin and Ryder, Nick and Subbiah, Melanie and Kaplan, Jared D and Dhariwal, Prafulla and Neelakantan, Arvind and Shyam, Pranav and Sastry, Girish and Askell, Amanda and others},
  journal={Advances in Neural Information Processing Systems},
  volume={33},
  pages={1877--1901},
  year={2020}
}

@article{openai2023gpt,
  title={{GPT-4 Technical Report}},
  author={OpenAI},
  journal={arXiv preprint arXiv:2303.08774},
  year={2023}
}

@inproceedings{reynolds2021prompt,
  title={{Prompt Programming for Large Language Models: Beyond the Few-Shot Paradigm}},
  author={Reynolds, Laria and McDonell, Kyle},
  booktitle={Extended Abstracts of the 2021 CHI Conference on Human Factors in Computing Systems},
  pages={1--7},
  year={2021}
}

@inproceedings{plaza-del-arco2023respectful,
  title={{Respectful or Toxic? Using Zero-Shot Learning with Language Models to Detect Hate Speech}},
  author={Plaza-del-arco, Flor Miriam and Nozza, Debora and Hovy, Dirk},
  booktitle={The 61st Annual Meeting Of The Association For Computational Linguistics},
  year={2023}
}

@article{bsharat2023principled,
  title={{Principled Instructions Are All You Need for Questioning LLaMA-1/2, GPT-3.5/4}},
  author={Bsharat, Sondos Mahmoud and Myrzakhan, Aidar and Shen, Zhiqiang},
  journal={arXiv preprint arXiv:2312.16171},
  year={2023}
}

@inproceedings{zhao2021calibrate,
  title={{Calibrate Before Use: Improving Few-shot Performance of Language Models}},
  author={Zhao, Zihao and Wallace, Eric and Feng, Shi and Klein, Dan and Singh, Sameer},
  booktitle={International Conference on Machine Learning},
  pages={12697--12706},
  year={2021},
}

@inproceedings{reif2022recipe,
  title={{A Recipe for Arbitrary Text Style Transfer with Large Language Models}},
  author={Reif, Emily and Ippolito, Daphne and Yuan, Ann and Coenen, Andy and Callison-Burch, Chris and Wei, Jason},
  booktitle={Proceedings of the 60th Annual Meeting of the Association for Computational Linguistics (Volume 2: Short Papers)},
  pages={837--848},
  year={2022}
}

@article{sahoo2024systematic,
  title={{A Systematic Survey of Prompt Engineering in Large Language Models: Techniques and Applications}},
  author={Sahoo, Pranab and Singh, Ayush Kumar and Saha, Sriparna and Jain, Vinija and Mondal, Samrat and Chadha, Aman},
  journal={arXiv preprint arXiv:2402.07927},
  year={2024}
}

@inproceedings{lu2022fantastically,
  title={{Fantastically Ordered Prompts and Where to Find Them: Overcoming Few-Shot Prompt Order Sensitivity}},
  author={Lu, Yao and Bartolo, Max and Moore, Alastair and Riedel, Sebastian and Stenetorp, Pontus},
  booktitle={Proceedings of the 60th Annual Meeting of the Association for Computational Linguistics (Volume 1: Long Papers)},
  pages={8086--8098},
  year={2022}
}

@article{wei2022chain,
  title={{Chain-of-Thought Prompting Elicits Reasoning in Large Language Models}},
  author={Wei, Jason and Wang, Xuezhi and Schuurmans, Dale and Bosma, Maarten and Xia, Fei and Chi, Ed and Le, Quoc V. and Zhou, Denny and others},
  journal={Advances in Neural Information Processing Systems},
  volume={35},
  pages={24824--24837},
  year={2022}
}

@article{kojima2022large,
  title={{Large Language Models are Zero-Shot Reasoners}},
  author={Kojima, Takeshi and Gu, Shixiang Shane and Reid, Machel and Matsuo, Yutaka and Iwasawa, Yusuke},
  journal={Advances in Neural Information Processing Systems},
  volume={35},
  pages={22199--22213},
  year={2022}
}

@inproceedings{chen2023mapo,
  title={{MAPO: Boosting Large Language Model Performance with Model-Adaptive Prompt Optimization}},
  author={Chen, Yuyan and Wen, Zhihao and Fan, Ge and Chen, Zhengyu and Wu, Wei and Liu, Dayiheng and Li, Zhixu and Liu, Bang and Xiao, Yanghua},
  booktitle={Findings of the Association for Computational Linguistics: EMNLP 2023},
  pages={3279--3304},
  year={2023}
}

@inproceedings{gonen2023demystifying,
    title = {{Demystifying Prompts in Language Models via Perplexity Estimation}},
    author = {Gonen, Hila  and
      Iyer, Srini  and
      Blevins, Terra  and
      Smith, Noah  and
      Zettlemoyer, Luke},
    booktitle = {Findings of the Association for Computational Linguistics: EMNLP 2023},
    year = {2023},
    address = {Singapore},
    pages = {10136--10148},
}

@misc{dairai2024leitfaden,
    title={{Leitfaden zum Prompt-Engineering}}, 
    author={DAIR.AI},
    year = {2024},
    note         = {Last accessed: 22.03.2024},
    howpublished = {\url{https://www.promptingguide.ai/de}}
}

@misc{openaiNDprompt,
    title={{Prompt Engineering}},
    author={OpenAI},
    year={n.D.},
    note         = {Last accessed: 22.03.2024},
    howpublished = {\url{https://platform.openai.com/docs/guides/prompt-engineering}}
}

@article{li2023large,
  title={{Large Language Models Understand and Can be Enhanced by Emotional Stimuli}},
  author={Li, Cheng and Wang, Jindong and Zhang, Yixuan and Zhu, Kaijie and Hou, Wenxin and Lian, Jianxun and Luo, Fang and Yang, Qiang and Xie, Xing},
  journal={arXiv preprint arXiv:2307.11760},
  year={2023}
}

@article{jin2022deep,
  title={{Deep Learning for Text Style Transfer: A Survey}},
  author={Jin, Di and Jin, Zhijing and Hu, Zhiting and Vechtomova, Olga and Mihalcea, Rada},
  journal={Computational Linguistics},
  volume={48},
  number={1},
  pages={155--205},
  year={2022},
  publisher={MIT Press One Broadway, 12th Floor, Cambridge, Massachusetts 02142, USA~…}
}

@article{hu2021causal,
  title={{A Causal Lens for Controllable Text Generation}},
  author={Hu, Zhiting and Li, Li Erran},
  journal={Advances in Neural Information Processing Systems},
  volume={34},
  pages={24941--24955},
  year={2021}
}

@article{zhang2023survey,
  title={{A Survey of Controllable Text Generation Using Transformer-Based Pre-Trained Language Models}},
  author={Zhang, Hanqing and Song, Haolin and Li, Shaoyu and Zhou, Ming and Song, Dawei},
  journal={ACM Computing Surveys},
  volume={56},
  number={3},
  pages={1--37},
  year={2023},
}

@inproceedings{prabhumoye2020exploring,
  title={{Exploring Controllable Text Generation Techniques}},
  author={Prabhumoye, Shrimai and Black, Alan W. and Salakhutdinov, Ruslan},
  booktitle={Proceedings of the 28th International Conference on Computational Linguistics},
  pages={1--14},
  year={2020}
}

@inproceedings{song2019generating,
  title={{Generating Responses with a Specific Emotion in Dialog}},
  author={Song, Zhenqiao and Zheng, Xiaoqing and Liu, Lu and Xu, Mu and Huang, Xuan-Jing},
  booktitle={Proceedings of the 57th Annual Meeting of the Association for Computational Linguistics},
  pages={3685--3695},
  year={2019}
}

@inproceedings{shen2020cdl,
  title={{CDL: Curriculum Dual Learning for Emotion-Controllable Response Generation}},
  author={Shen, Lei and Feng, Yang},
  booktitle={Proceedings of the 58th Annual Meeting of the Association for Computational Linguistics},
  pages={556--566},
  year={2020}
}

@inproceedings{colombo2019affect,
  title={{Affect-Driven Dialog Generation}},
  author={Colombo, Pierre and Witon, Wojciech and Modi, Ashutosh and Kennedy, James and Kapadia, Mubbasir},
  booktitle={Proceedings of the 2019 Conference of the North American Chapter of the Association for Computational Linguistics: Human Language Technologies, Volume 1 (Long and Short Papers)},
  pages={3734--3743},
  year={2019}
}

@inproceedings{resendiz2023emotion,
  title={{Emotion-Conditioned Text Generation through Automatic Prompt Optimization}},
  author={Resendiz, Yarik Menchaca and Klinger, Roman},
  booktitle={Proceedings of the 1st Workshop on Taming Large Language Models: Controllability in the Era of Interactive Assistants},
  pages={24--30},
  year={2023}
}

@article{coda-forno2023inducing,
  title={{Inducing Anxiety in Large Language Models Increases Exploration and Bias}},
  author={Coda-Forno, Julian and Witte, Kristin and Jagadish, Akshay K. and Binz, Marcel and Akata, Zeynep and Schulz, Eric},
  journal={arXiv preprint arXiv:2304.11111},
  year={2023}
}

@misc{openaiNDfinetuning,
    title={{Fine-Tuning}}, 
    author={OpenAI},
    year={n.D.},
    note         = {Last accessed: 29.04.2024},
    howpublished = {\url{https://platform.openai.com/docs/guides/fine-tuning}}
}

@inproceedings{demszky2020goemotions,
  title={{GoEmotions: A Dataset of Fine-Grained Emotions}},
  author={Demszky, Dorottya and Movshovitz-Attias, Dana and Ko, Jeongwoo and Cowen, Alan and Nemade, Gaurav and Ravi, Sujith},
  booktitle={Proceedings of the 58th Annual Meeting of the Association for Computational Linguistics},
  pages={4040--4054},
  year={2020}
}

@inproceedings{poria2019meld,
  title={{MELD: A Multimodal Multi-Party Dataset for Emotion Recognition in Conversations}},
  author={Poria, Soujanya and Hazarika, Devamanyu and Majumder, Navonil and Naik, Gautam and Cambria, Erik and Mihalcea, Rada},
  booktitle={Proceedings of the 57th Annual Meeting of the Association for Computational Linguistics},
  pages={527--536},
  year={2019}
}

@misc{swisscenterforaffectivesciencesNDresearch,
    title={{Research Material and Online Research}}, 
    author={{Swiss Center for Affective Sciences}},
    year={n.D.},
    note         = {Last accessed: 16.04.2024},
    howpublished = {\url{https://www.unige.ch/cisa/research/materials-and-online-research/research-material/}}
}

@article{mcrae2008gender,
  title={{Gender Differences in Emotion Regulation: An fMRI Study of Cognitive Reappraisal}},
  author={McRae, Kateri and Ochsner, Kevin N. and Mauss, Iris B. and Gabrieli, John J.D. and Gross, James J.},
  journal={Group Processes \& Intergroup Relations},
  volume={11},
  number={2},
  pages={143--162},
  year={2008}
}

@article{brody1997gender,
  title={{Gender and Emotion: Beyond Stereotypes}},
  author={Brody, Leslie R.},
  journal={Journal of Social issues},
  volume={53},
  number={2},
  pages={369--393},
  year={1997},
  publisher={Wiley Online Library}
}

@inproceedings{sheng2021societal,
  title={{Societal Biases in Language Generation: Progress and Challenges}},
  author={Sheng, Emily and Chang, Kai-Wei and Natarajan, Prem and Peng, Nanyun},
  booktitle={Proceedings of the 59th Annual Meeting of the Association for Computational Linguistics and the 11th International Joint Conference on Natural Language Processing (Volume 1: Long Papers)},
  pages={4275--4293},
  year={2021}
}

@article{mohammad2013crowdsourcing,
  title={{Crowdsourcing a Word-Emotion Association Lexicon}},
  author={Mohammad, Saif M. and Turney, Peter D.},
  journal={Computational Intelligence},
  volume={29},
  number={3},
  pages={436--465},
  year={2013},
}

@inproceedings{mohammad2010emotions,
  title={{Emotions Evoked by Common Words and Phrases: Using Mechanical Turk to Create an Emotion Lexicon}},
  author={Mohammad, Saif and Turney, Peter},
  booktitle={Proceedings of the NAACL HLT 2010 Workshop on Computational Approaches to Analysis and Generation of Emotion in Text},
  pages={26--34},
  year={2010}
}

@inproceedings{zhang2019bertscore,
  title={{BERTScore: Evaluating Text Generation with BERT}},
  author={Zhang, Tianyi and Kishore, Varsha and Wu, Felix and Weinberger, Kilian Q. and Artzi, Yoav},
  booktitle={International Conference on Learning Representations},
  year={2019}
}

@misc{huggingfaceNDbert_score,
    title={Metric: bert\_score},
    author={{Hugging Face}},
    year={n.D.},
    note={Last accessed: 21.05.2024},
    howpublished={\url{https://huggingface.co/spaces/evaluate-metric/bertscore}}
}

@article{flesch1948new,
  title={{A New Readability Yardstick}},
  author={Flesch, Rudolph},
  journal={Journal of Applied Psychology},
  volume={32},
  number={3},
  pages={221},
  year={1948},
}

@misc{fleschNDhow,
    title={{How to Write Plain English}},
    author={Flesch, Rudolph},
    year={n.D.},
    note={Last accessed: 26.05.2024},
    howpublished={\url{https://web.archive.org/web/20160712094308/http://www.mang.canterbury.ac.nz/writing_guide/writing/flesch.shtml}}
}

@inproceedings{lindquist2016language,
    author={Kristen A. Lindquist and Maria Gendron and Ajay B. Satpute},
    title={{Language and Emotion: Putting Words into Feelings and Feelings into Words}},
    booktitle={Handbook of Emotions},
    editor={Lewis, Michael and Haviland-Jones, Jeannette M and Feldman Barrett, Lisa},
    year={2016},
    publisher={Guilford Press},
    pages = {579--594},
}

@inproceedings{curry2023computer,
  title={{Computer Says "No": The Case Against Empathetic Conversational AI}},
  author={Curry, Alba and Cercas Curry, Amanda},
  booktitle={Findings of the Association of Computer Linguistics: ACL 2023},
  year={2023},
  organization={Association for Computational Linguistics}
}

@inproceedings{deshpande2023anthropomorphization,
  title={{Anthropomorphization of AI: Opportunities and Risks}},
  author={Deshpande, Ameet and Rajpurohit, Tanmay and Narasimhan, Karthik and Kalyan, Ashwin},
  booktitle={Proceedings of the Natural Legal Language Processing Workshop 2023},
  pages={1--7},
  year={2023}
}

@article{giger2019humanization,
  title={{Humanization of Robots: Is It Really Such a Good Idea?}},
  author={Giger, Jean-Christophe and Pi{\c{c}}arra, Nuno and Alves-Oliveira, Patr{\'\i}cia and Oliveira, Raquel and Arriaga, Patr{\'\i}cia},
  journal={Human Behavior and Emerging Technologies},
  volume={1},
  number={2},
  pages={111--123},
  year={2019},
  publisher={Wiley Online Library}
}

@article{ryan2020ai,
  title={{In AI We Trust: Ethics, Artificial Intelligence, and Reliability}},
  author={Ryan, Mark},
  journal={Science and Engineering Ethics},
  volume={26},
  number={5},
  pages={2749--2767},
  year={2020},
  publisher={Springer}
}

@misc{statista2023chatbots,
    title={{Chatbot-Revolution durch ChatGPT}}, 
    author={Statista},
    year={2023},
    note         = {Last accessed: 11.12.2024},
    howpublished = {\url{https://de.statista.com/statistik/studie/id/134940/dokument/chatbot-revolution-durch-chatgpt/}}
}

@inproceedings{ghandeharioun2019understanding,
  title={{Towards Understanding Emotional Intelligence for Behavior Change Chatbots}},
  author={Ghandeharioun, Asma and McDuff, Daniel and Czerwinski, Mary and Rowan, Kael},
  booktitle={2019 8th International Conference on Affective Computing and Intelligent Interaction (ACII)},
  pages={8--14},
  year={2019},
}

@article{fitzpatrick2017delivering,
  title={{Delivering Cognitive Behavior Therapy to Young Adults With Symptoms of Depression and Anxiety Using a Fully Automated Conversational Agent (Woebot): A Randomized Controlled Trial}},
  author={Fitzpatrick, Kathleen Kara and Darcy, Alison and Vierhile, Molly},
  journal={JMIR Mental Health},
  volume={4},
  number={2},
  pages={e7785},
  year={2017},
  publisher={JMIR Publications Inc., Toronto, Canada}
}

@inproceedings{rodriguez2023qualitative,
  title={{Qualitative Analysis of Conversational Chatbots to Alleviate Loneliness in Older Adults as a Strategy for Emotional Health}},
  author={Rodr{\'\i}guez-Mart{\'\i}nez, Antonia and Amezcua-Aguilar, Teresa and Cort{\'e}s-Moreno, Javier and Jim{\'e}nez-Delgado, Juan Jos{\'e}},
  booktitle={Healthcare},
  volume={12},
  pages={62},
  year={2023},
}

@article{yun2022effects,
  title={{The Effects of Chatbot Service Recovery with Emotion Words on Customer Satisfaction, Repurchase Intention, and Positive Word-of-Mouth}},
  author={Yun, Jeewoo and Park, Jungkun},
  journal={{Frontiers in Psychology}},
  volume={13},
  pages={922503},
  year={2022},
  publisher={Frontiers Media SA}
}

@article{kim2007pedagogical,
  title={{Pedagogical Agents as Learning Companions: The Impact of Agent Emotion and Gender}},
  author={Kim, Yanghee and Baylor, Amy L and Shen, Entong},
  journal={Journal of Computer Assisted Learning},
  volume={23},
  number={3},
  pages={220--234},
  year={2007},
  publisher={Wiley Online Library}
}

\newpage


\end{document}